\documentclass[letterpaper]{article} % DO NOT CHANGE THIS
\usepackage{aaai24}  % DO NOT CHANGE THIS
\usepackage{times}  % DO NOT CHANGE THIS
\usepackage{helvet}  % DO NOT CHANGE THIS
\usepackage{courier}  % DO NOT CHANGE THIS
\usepackage[hyphens]{url}  % DO NOT CHANGE THIS
\usepackage{graphicx} % DO NOT CHANGE THIS
\usepackage{natbib}  % DO NOT CHANGE THIS AND DO NOT ADD ANY OPTIONS TO IT
\usepackage{caption} % DO NOT CHANGE THIS AND DO NOT ADD ANY OPTIONS TO IT
\usepackage{algorithm}
\usepackage{algorithmic}
\usepackage{amsmath}
\usepackage{multicol}
\usepackage{multirow}
\usepackage{pifont}
\usepackage{subcaption}
\usepackage{newfloat}
\usepackage{listings}
\DeclareCaptionStyle{ruled}{labelfont=normalfont,labelsep=colon,strut=off} % DO NOT CHANGE THIS
\floatstyle{ruled}
\newfloat{listing}{tb}{lst}{}
\floatname{listing}{Listing}
\title{Mean Teacher DETR with Masked Feature Alignment: \\
A Robust Domain Adaptive Detection Transformer Framework}
\author{
    Weixi Weng,
    Chun Yuan
}
\affiliations{
    Tsinghua Shenzhen International Graduate School, Tsinghua University\\
    wengwx22@mails.tsinghua.edu.cn, yuanc@sz.tsinghua.edu.cn
}

\usepackage{bibentry}
\begin{document}

\maketitle

\begin{abstract}
Unsupervised domain adaptive object detection (UDAOD) research on Detection Transformer (DETR) mainly focuses on feature alignment and existing methods can be divided into two kinds, each of which has its unresolved issues. One-stage feature alignment methods can easily lead to performance fluctuation and training stagnation. Two-stage feature alignment method based on mean teacher comprises a pretraining stage followed by a self-training stage, each facing problems in obtaining a reliable pretrained model and achieving consistent performance gains. Methods mentioned above have not yet explored how to utilize the third related domain such as the target-like domain to assist adaptation. To address these issues, we propose a two-stage framework named MTM, \emph{i.e.} \textbf{M}ean \textbf{T}eacher-DETR with \textbf{M}asked Feature Alignment. In the pretraining stage, we utilize labeled target-like images produced by image style transfer to avoid performance fluctuation. In the self-training stage, we leverage unlabeled target images by pseudo labels based on mean teacher and propose a module called Object Queries Knowledge Transfer (OQKT) to ensure consistent performance gains of the student model. Most importantly, we propose masked feature alignment methods including Masked Domain Query-based Feature Alignment (MDQFA) and Masked Token-Wise Feature Alignment (MTWFA) to alleviate domain shift in a more robust way, which not only prevent training stagnation and lead to a robust pretrained model in the pretraining stage but also enhance the model's target performance in the self-training stage. Experiments on three challenging scenarios and a theoretical analysis verify the effectiveness of MTM.
\end{abstract}

\section{Introduction}

Object detection is always recognized as an important task in the field of computer vision. Recently, 
Detection Transformer (DETR) \cite{carion2020detr} redefines object prediction by departing from the traditional anchor-based methodology \cite{girshick2015fast} and embracing the concept of object queries which serve as learnable embeddings interacting with the image features, allowing the model to predict object classes and bounding box coordinates. While DETR has demonstrated remarkable performance on various datasets, its application to real-world environments still presents challenges, particularly when training data and testing data are collected from different distributions, \emph{i.e.} domain shift.

\begin{figure*}[t]
\centering
    \includegraphics[width=0.90\textwidth]{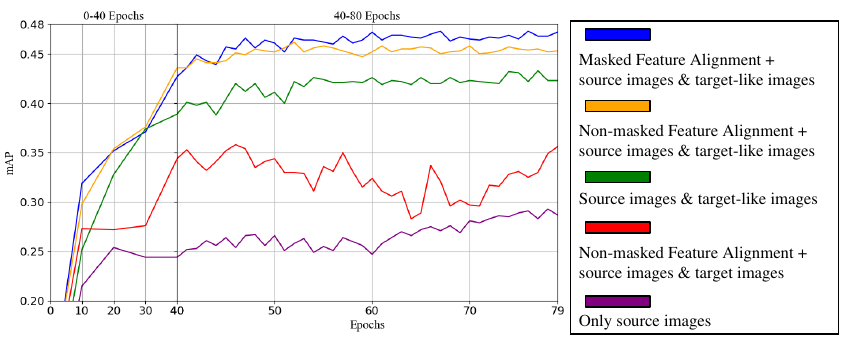}
    \caption{We train Deformable DETR from scratch with different pretraining strategies in the weather adaptation scenario. It turns out that pretraining with masked feature alignment on source images and target-like images brings the best performance.}
    \label{fig:performance curves}
\end{figure*}
% 说清楚这个图干嘛的
% 为什么可以解决这个问题，要说清楚
In order to enable object detectors trained on a labeled source domain to be effectively deployed on a completely unlabeled target domain, unsupervised domain adaptive object detection(UDAOD) emerged as a solution. 
UDAOD research on DETR is gradually gaining increasing attention, with particular emphasis on feature alignment \cite{wang2021exploring,gong2022improving,yu2022cross,zhang2023detr}. Feature alignment is commonly implemented through adversarial training, with the objective of extracting domain-invariant global-level, local-level and instance-level features from source domain and target domain, thus reducing domain shift. Existing methods on DETR can be categorized into one-stage feature alignment methods \cite{wang2021exploring,gong2022improving,zhang2023detr} and two-stage feature alignment method based on mean teacher \cite{yu2022cross}. 
We deeply studied the above two methods and identified several inherent issues, which will be elaborated below.

One-stage feature alignment methods \cite{xu2020exploring,gong2022improving,zhang2023detr} generally train DETR from scratch with labeled source images and unlabeled target images, conducting feature alignment on the output of the backbone, encoder, and decoder. However, training in this manner has inherent issues:
% fluctuation and low quality
(I1) performance fluctuation: the lack of labels for target images restricts the model's ability to extract features from target images, and conducting feature alignment with low-quality target features easily results in performance fluctuation. Furthermore, it's very likely to end up with an underperforming model as shown by the red line in Figure \ref{fig:performance curves}; (I2) training stagnation: 
at the early training stage, if there is a significant difference in the distribution between the two types of training data, domain discriminators converge quickly and become proficient at distinguishing features in the later period of training. However, the backbone struggles to produce domain-invariant features at this point, which results in marginal performance improvement as shown by the yellow line in Figure \ref{fig:performance curves}.

% 紧接着就说这个方法的好处
Two-stage feature alignment method \cite{yu2022cross} based on mean teacher \cite{sohn2020simple,xu2021end} aims to leverage unlabeled target images by self-training.
Take MTTrans \cite{yu2022cross} for example, it comprises a pretraining stage followed by a self-training stage. It firstly pretrains a model merely on labeled source images and only conducts feature alignment in the self-training stage, sharing the same issues with the one-stage methods.
Although MTTrans significantly improves target performance through mean teacher, it also faces other issues: (I3) unreliable pretrained model: its pretrained model is obtained by training on labeled source images without any feature alignment, resulting in the teacher model failing to generate accurate pseudo-labels; (I4) unstable performance gains: it uses the same the object queries between the teacher and the student model in the self-training stage, resulting in the student model struggling to attain consistent performance gains.

Furthermore, the methods mentioned above only consider adapting between two domains, while how to use the third related domain such as a target-like domain generated by image style transfer to assist adaptation is still unexplored.

% 为什么解决了这个问题
Based on the facts above, we propose MTM, \emph{i.e.} \textbf{M}ean \textbf{T}eacher-DETR with \textbf{M}asked Feature Alignment consisting of a pretraining stage and a self-training stage.
In the pretraining stage, we utilize labeled source images and labeled target-like images produced by cycleGAN \cite{zhu2017unpaired} instead of unlabeld target images to avoid the problem of performance fluctuation (I1). 
% Target-like images retains the content of the source image but exhibits the style of the target domain.
In the self-training stage, we adopt the mean teacher framework to leverage target images by pseudo labels. We further propose Object Queries Knowledge Transfer (OQKT) which enhances the semantic information of the student model's object queries by multi-head attention to prompt consistent performance gains of the student model (I4). Most importantly, we propose masked feature alignment methods including Masked Domain Query-based Feature Alignment (MDQFA) and Masked Token-Wise Feature Alignment (MTWFA) to alleviate domain shift in a more robust way, which not only prevent training stagnation (I2) and lead to robust pretrained model (I3) in the pretraining stage but also enhance the model's final target performance in the self-training stage.
The contributions of this paper are as follows:

\begin{itemize}
\item Through experiments we find that utilizing labeled target-like images produced by CycleGAN \cite{zhu2017unpaired} to participate in training DETR from scratch avoids performance fluctuation.
% 不能用prove
\item We propose Object Queries Knowledge Transfer (OQKT) based on mean teacher DETR to guarantee consistent performance gains of the student model. 
\item Most importantly, we propose masked feature alignment methods including Masked Domain Query-based Feature Alignment (MDQFA) and Masked Token-Wise Feature Alignment (MTWFA) to reduce domain shift in a more robust way. They benefit both stages of MTM.
\item Extensive experiments on three challenging domain adaptation scenarios have demonstrated that MTM has outperformed existing state-of-the-art (SOTA) methods in this field. A theoretical analysis is also presented to verify the effectiveness of MTM.

\end{itemize}

\section{Related Work}
\subsection{Detection Transformer}
Object detection is a challenging task in the field of computer vision. Recently, the Detection Transformer(DETR) has garnered significant attention by presenting a novel object detection pipeline based on Transformer.
% 加一段更详细的从编码、oq到解码的介绍
Deformable DETR \cite{zhu2021deformable}, as an important advancement of DETR, introduces deformable attention modules that only attend to a small set of key sampling points around a reference. Deformable DETR achieves better performance than DETR with much less convergence time and computational requirements. Deformable DETR has several variants and the two-stage variant achieves the best performance. In the two-stage Deformable DETR, the encoder generates object queries from multi-scale features extracted by the CNN backbone, while the decoder refines object queries and employs them for predicting object classes and generating bounding box coordinates. Following \cite{wang2021exploring}, we mainly investigate UDAOD on Deformable DETR, specifically the two-stage Deformable DETR.

\subsection{Domain Adaptive Object Detection}
Extensive research has been conducted on unsupervised domain adaptive object detection (UDAOD) based on different architectures. Proposed solutions can be categorized into three mainstreams: feature alignment, mean teacher and domain transfer. These kinds of solutions often work together to form a more robust framework. However, research on DETR so far has mainly focused on feature alignment.

DAF \cite{chen2018domain} is the first work to apply feature alignment on Faster RCNN by means of adversarial training, performing feature alignment at the image-level and instance-level respectively.
Recently, there has been a continuous surge of research based on DETR. SFA \cite{wang2021exploring} makes a pioneering attempt to conduct sequence feature alignment on different levels based on DETR. $\text{O}^2$Net \cite{gong2022improving} introduces an Object-Aware Alignment module to align the multi-scale features of the CNN backbone. DA-DETR \cite{zhang2023detr} proposes a CNN-Transformer Blender which fuses the output of the CNN backbone and the encoder to better align them. 

MTOR \cite{cai2019exploring} firstly exploits mean teacher on Faster RCNN, building a consistency loss between the teacher and the student model to learn more about the object relation. MTTrans \cite{yu2022cross} firstly applies mean teacher on DETR, proposing multi-level feature alignment to improve the quality of pseudo labels. 

Methods of domain transfer are normally built on CycleGAN \cite{zhu2017unpaired} which conducts image style transfer between domains. Given source images and target images, CycleGAN generates target-like images by transferring source images into target style, as well as source-like images. UMT \cite{deng2021unbiased} additionally utilizes source-like and target-like images for training, and proposes several strategies on mean teacher to address model bias. AWADA \cite{menke2022awada} uses the proposals generated by the object detector to aid in more efficient image style transfer. While domain transfer has been extensively studied on other object detectors \cite{ren2015faster}, to the best of our knowledge, how to use it to assist domain adaptation on DETR remains unexplored.

\begin{figure*}[ht]
\centering
    \includegraphics[width=0.90\textwidth]{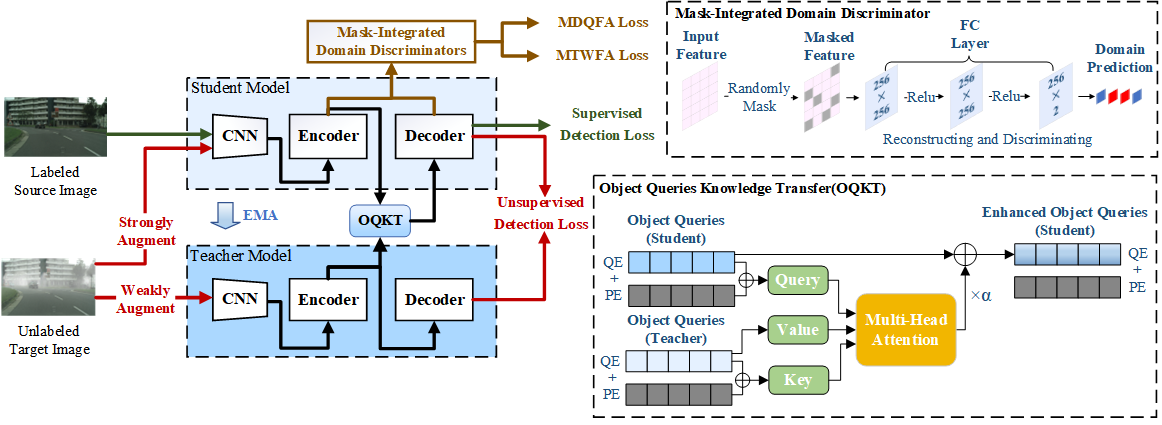}
    \caption{MTM framework in the self-training stage. In the mean teacher structure, the student model is updated by back-propagation while the teacher model is updated by the Exponential Moving Average (EMA) of the student model. OQKT enhances the student's object queries with the teacher's object queries by the multi-head attention mechanism. Masked feature alignment is also conducted in this stage to alleviate the domain shift between source data and target data.}
\label{fig:framework}
\end{figure*}

\section{Methods}
This section introduces MTM consisting of a robust pretraining stage and a performance-enhancing self-training stage.

\subsection{MTM Framework Overview}
In the pretraining stage, we utilize labeled source images and labeled target-like images for training. A target-like image produced by CycleGAN \cite{zhu2017unpaired} retains the content of the source image while exhibiting the style of target images, thereby sharing identical annotations with its corresponding source images. Pretraining with labeled target-like images not only avoids performance fluctuation (I1) as shown by the green line in Figure \ref{fig:performance curves} but also contributes to obtaining a robust pretrained model (I3).

In the self-training stage, we utilize labeled source images and unlabeled target images for training. Given the absence of target images during pretraining, we leverage them by pseudo labels based on mean teacher in this stage. We further propose Object Queries Knowledge Transfer (OQKT) based on mean teacher DETR to ensure consistent performance gains of the student model (I4). 

Most importantly, we propose masked feature alignment methods including Masked Domain Query-based Feature Alignment (MDQFA) and Masked Token-Wise Feature Alignment (MTWFA) which not only prevent training stagnation (I2) and help to obtain a robust pretrained model (I3) in the pretraining stage but also contribute to enhancing the model's final target performance in the self-training stage.

\subsection{Object Queries Knowledge Transfer}\label{sec:OQKT}
In the mean teacher framework, the teacher model takes in weakly augmented target images as input while the student model processes strongly augmented counterparts. As they undergo different data augmentation strategies, it is highly likely to result in significantly different object queries between them. Therefore, enforcing the student to use the same object queries as the teacher \cite{yu2022cross} tends to restrain consistent performance gains of the student. 
However, we can't ignore that the teacher model is likely to generate valuable object queries from weakly augmented target images. In order to transfer the knowledge within the object queries from the teacher model to the student model and address the issue of unstable performance gains (I4), we propose a simple yet effective module named \textbf{O}bject \textbf{Q}ueries \textbf{K}nowledge \textbf{T}ransfer(OQKT).

$QE_t$ represents query embeddings of the teacher model's object queries and $PE_t$ is the corresponding position embeddings. The teacher model's object queries are obtained by adding $QE_t$ and $PE_t$ together. In a similar way, $QE_s$ and $PE_s$ are defined. We further define:
\begin{equation}
\fontsize{9}{10}\selectfont
\begin{aligned}
    &Query = QE_{s} + PE_{s} \\
    &Key = QE_{t} + PE_{t} \\
    &Value = QE_{t} \\
\end{aligned}
\end{equation}

The knowledge transfer mechanism of this module is based on the multi-head attention mechanism in \cite{vaswani2017attention}, which is shown as follows:
\begin{equation}
\fontsize{9}{10}\selectfont
    \textsc{Attention}(Q,K,V) = \textup{softmax}(\frac{Q K^{T}}{\sqrt{d_{k}}}V)
\end{equation}
\begin{equation}
\fontsize{9}{10}\selectfont
\begin{split}
    \textsc{MultiHead}(Q,K,V) = \textup{Concat}(head_1, \cdots head_n)W^{O} \\
    where \ head_{i} = \textsc{Attention}(QW^{Q}_{i},KW^{K}_{i},VW^{V}_{i})
\end{split}
\end{equation}

Where $W^{Q}_{i}$,$W^{K}_{i} $, $W^{V}_{i}$ and $W^{O}_{i}$ are weight matrices and $n=16, d_k=16$ in OQKT. The teacher's object queries and the student's object queries interact via the multi-head attention mechanism to obtain additional features. The additional features are multiplied by a parameter and added to the student's query embeddings, enhancing the semantic information of the student's object queries for better prediction. $QE_S$ is updated as follows:
\begin{equation}
\fontsize{9}{10}\selectfont
    QE_S = QE_S + \alpha \times \textsc{MultiHead}(Query,Key,Value)
\end{equation}

In the self-training stage, with the increasing of epochs, $\alpha$ decreases linearly from 1 to 0, promoting the student model to generate high-quality object queries by itself.

\subsection{Masked Feature Alignment}
Proposed by \cite{wang2021exploring}, Domain Query-based Feature Alignment (DQFA) alleviates global-level domain shift from image style, weather \emph{etc.,} while Token-Wise Feature Alignment (TWFA) deals with local-level and instance-level domain shift caused by object appearance, scale, texture \emph{etc}. However, when conducting them on source images and target-like images in the pretraining stage, training stagnation is observed as shown by the yellow line in Figure \ref{fig:performance curves}.

To prevent training stagnation and elevate the model's target performance, we innovatively propose masked feature alignment methods including Masked Domain Query-based Feature Alignment (MDQFA) and Masked Token-Wise Feature Alignment (MTWFA). MDQFA and MTWFA randomly mask sequence features before feeding them into domain discriminators, making it harder to classify their domains. Our masked feature alignment methods not only hinder the domain discriminators from converging too quickly thus preventing training stagnation, but also improve the domain discriminators' robustness through a diverse range of masked sequence features, consequently further elevating the model's target performance. Details of MDQFA and MTWFA are described below. The domain discriminators' structures for both masked feature alignment methods exhibit identical architectures, as depicted in Figure \ref{fig:framework}.

\subsubsection{Masked Domain Query-based Feature Alignment}

On the encoder side, a domain query $q^{enc} \in R^{1 \times C}$ is concatenated with the multi-scale feature extracted from the CNN backbone to form the input for the encoder:
\begin{equation}
\fontsize{9}{10}\selectfont
    Z_{0} = [q^{enc}, z_{1}, z_{2}, ...,z_{N_{enc}}] + E_{pos} + E_{level}
\end{equation}
$N_{enc}$ stands for the length of the encoder input embeddings. $E_{pos}$ and $E_{level}$ represents corresponding position embeddings and feature level embeddings. The sequence features derived from the $l$-th layer are denoted as $Z_l$. During the encoding process, the domain query gathers domain-specific features from the sequence feature. We further set a random mask which can be formulated as:
\begin{equation}
\fontsize{9}{10}\selectfont
    M^{\text{DQFA}}_{enc_{l,i}} = 
        \left\{
            \begin{array}{ll}
               0,  &  \text{if$\,$} R_{l,i} < \theta_{Mask}\\
               1,  &  \text{Otherwise}
            \end{array} 
        \right.
\end{equation}
where $l=1...L_{enc}$ indexes the layer of the encoder, $i\in [0, C)$ represents the element coordinate of domain query, $R_{l, i}$ is a random floating-point number that falls within the range of 0 to 1 and $\theta_{Mask}$ is a hyperparameter controlling the masking rate. Then we compute the element-wise product of the domain query and the mask to get the masked domain query $Z_{l,0}^M$ that only contains partial global domain information. The domain discriminator $D^{\text{MDQFA}}_{enc}$ needs to determine which domain the masked domain query belongs to. We utilize the binary cross-entropy loss as the domain classification loss. The whole procedure is formulated as below:
\begin{equation}
\fontsize{9}{10}\selectfont
    Z_{l,0}^M = M^{\text{DQFA}}_{enc_{l}} \odot Z_{l,0}
\label{eq: MDQFA}    
\end{equation}
\begin{equation}    
\fontsize{9}{10}\selectfont
\begin{aligned}
    \mathcal{L}_{enc_{l}}^{\text{MDQFA}} &=-\bigl[dlogD^{\text{MDQFA}}_{enc}(Z^{M}_{l,0}) \\ 
    &+ (1-d)log\bigl(1-D^{\text{MDQFA}}_{enc}(Z^{M}_{l,0})\bigr)\bigr]
\end{aligned}
\label{eq: MDQFA loss}
\end{equation}

$d$ represents the domain label with 0 denoting the source domain and 1 denoting the target or target-like domain.

On the decoder side, we concatenate a domain query $q^{dec} \in R^{1 \times C}$ with the object queries to form the input: 
\begin{equation}
    Q_{0} = [q^{dec};q_{1},q_{2},...,q_{N_{dec}}]+E^{'}_{pos} 
\end{equation}
$N_{dec}$ stands for the length of the decoder input embeddings. $E^{'}_{pos}$ represents position embeddings. The refined object queries derived from the $l$-th layer are denoted as $Q_l$. The generation of the random mask of the domain queries in the decoder $M^{\text{DQFA}}_{dec_{l}}$ follows a similar procedure as equation \ref{eq: MDQFA}.
We feed the masked domain query into the domain discriminator $D_{dec}^{\text{MDQFA}}$ and calculate the binary cross-entropy loss.

\subsubsection{Masked Token-Wise Feature Alignment}

On the encoder side, each token embedding in the sequence feature stands for a particular-scale feature of the image and aggregates local-level information from nearby key points. We set a random mask for the whole sequence feature as follows:
\begin{equation}
\fontsize{9}{10}\selectfont
    M^{\text{TWFA}}_{enc_{l,i,j}} = 
        \left\{
            \begin{array}{ll}
               0,  &  \text{if$\,$} R_{l,i,j} < \eta  \cdot \theta_{Mask}\\ 
               1,  &  \text{Otherwise}
            \end{array} 
        \right.
\label{eq: MTWFA}
\end{equation}
where $l = 1...L_{dec}$ indexes the layers of the encoder, $i \in [1,N_{enc}]$ and $j \in [0,C)$ respectively represent the horizontal and vertical coordinates of the sequence feature. Since domain discriminators of MTWFA deal with more diverse local-level features which are more difficult to classify their domains, we set the hyperparameter $\eta$ to a decimal between 0 and 1 indicating that MTWFA will use a smaller mask threshold compared with MDQFA.
Then we compute the element-wise product of the sequence feature and the corresponding mask as the masked sequence feature, which is fed into the domain discriminator $D^{\text{MTWFA}}_{enc}$. We adopt the binary cross-entropy loss as the domain classification loss. The whole procedure is formulated as below:
\begin{equation}
\fontsize{9}{10}\selectfont
    Z_{l}^M = M^{\text{TWFA}}_{enc_{l}} \odot Z_{l}
\end{equation}
\begin{equation}
\fontsize{9}{10}\selectfont
\begin{aligned}
    \mathcal{L}^{\text{MTWFA}}_{enc_l}=-\frac{1}{N_{enc}}\sum^{N_{enc}}_{i=1}\bigl[dlog D^{\text{MTWFA}}_{enc}(Z^{M}_{l,i})  \\
    + (1-d)log\bigl(1-D^{\text{MTWFA}}_{enc}(Z^{M}_{l,i})\bigr)\bigr]
\end{aligned}
\label{eq: MTWFA loss}
\end{equation}

$d$ signifies the domain label, where 0 denotes the source domain, and 1 denotes the target or target-like domain.

On the decoder side, each object query that corresponds to a predicted object's class and position aggregates instance-level features in the decoding process. Similarly, the sequence features of the decoder are randomly masked and then fed into the domain discriminator $D^{\text{MTWFA}}_{dec}$ to calculate the corresponding binary cross-entropy loss.
\begin{table*}[htb]
\fontsize{9pt}{10pt}\selectfont
\centering    
    \begin{tabular}{c|c c c c c c c c|c}
        \hline
        Method & person & rider & car & truck & bus & train &motorcycle & bicycle & mAP \\ 
        \hline
        Deformable DETR(source) & 38.0 & 38.7 & 45.3 & 16.3 & 26.7 & 4.2 & 22.9 & 36.7 & 28.5 \\
        SFA\cite{wang2021exploring} & 46.5 & 48.6 & 62.6 & 25.1 & 46.2 & 29.4 & 28.3 & 44.0 & 41.3 \\
        MTTrans\cite{yu2022cross} & 47.7 & 49.9 & 65.2 & 25.8 & 45.9 & 33.8 & 32.6 & 46.5 & 43.4 \\
        DA-DETR\cite{zhang2023detr} & 49.9 & 50.0 & 63.1 & 25.8 & 45.9 & 33.8 & 32.6 & 46.5 & 43.5 \\
        $\text{O}^{2}$Net\cite{gong2022improving} & 48.7 & 51.5 & 63.6 & 31.1 & 47.6 & \textbf{47.8} & 38.0 & 45.9 & 46.8 \\
        MTM(ours) & \textbf{51.0} & \textbf{53.4} & \textbf{67.2} & \textbf{37.2} & \textbf{54.4} & 41.6 & \textbf{38.4} & \textbf{47.7} & \textbf{48.9} \\
        \hline
    \end{tabular}
    \caption{Results(\%) in the weather adaptation scenario, \emph{i.e.} Cityscapes $\rightarrow$ Foggy Cityscapes.}
    \label{tab:weather results}
\end{table*}

\begin{table*}[htb]
\fontsize{9pt}{10pt}\selectfont
\centering
    \begin{tabular}{c|c c c c c c c|c}
        \hline
        Method & person & rider & car & truck & bus & motorcycle & bicycle & mAP \\ 
        \hline
        % \hline
        % Faster RCNN(source) & Faster RCNN & 28.8 & 25.4 & 44.1 & 17.9 & 16.1 & 13.9 & 22.4 & 24.1 \\
        % DAF & Faster RCNN & 28.9 & 27.4 & 44.2 & 19.1 & 18.0 & 14.2 & 22.4 & 24.9 \\
        % SW-DA & Faster RCNN & 29.5 & 29.9 & 44.8 & 20.2 & 20.7 & 15.2 & 23.1 & 26.2 \\
        % AWADA & Faster RCNN &  41.5 & 34.2 & 56.0 & 18.7 & 20.0 & 20.4 & 29.7 & 31.5\\ 
        % \hline
        Deformable DETR(source) & 38.9 & 26.7 & 55.2 & 15.7 & 19.7 & 10.8 & 16.2 & 26.2 \\
        SFA\cite{wang2021exploring} & 40.4 & 27.6 & 57.5 & 19.1 & 23.4 & 15.4 & 19.2 & 28.9 \\
        $\text{O}^{2}$Net\cite{gong2022improving} & 40.4 & 31.2 & 58.6 & 20.4 & 25.0 & 14.9 & 22.7 & 30.5 \\
        MTTrans\cite{yu2022cross} & 44.1 & 30.1 & 61.5 & 25.1 & 26.9 & 17.7 & 23.0 & 32.6\\
        MTM(ours) & \textbf{53.7} &  \textbf{35.1} & \textbf{68.8} & \textbf{23.0} & \textbf{28.8} & \textbf{23.8} & \textbf{28.0} & \textbf{37.3} \\
        \hline
    \end{tabular}
    \caption{Results(\%) in the scene adaptation scenario, \emph{i.e.} Cityscapes $\rightarrow$ BDD100K.}
    \label{tab:scene results}
\end{table*}

\subsection{Overall Training Strategy}
Our MTM framework consists of two training stages. In the pretraining stage, we aim to train a robust pretrained model, which will serve as the teacher model later. The loss function of this stage $L_{pre}$ combines the object detection loss and the feature alignment loss, which is defined as follows:
\begin{equation}
\fontsize{9}{10}\selectfont
    \mathcal{L}_{pre} = \mathcal{L}_{det}(I_{s},B_{s})+ \mathcal{L}_{det}(I_{tl},B_{s}) - \mathcal{L}_{adv}(I_{s}) - \mathcal{L}_{adv}(I_{tl})
\end{equation}

Where $I_{s}$ and $I_{tl}$ represent source images and target-like images, and they share the same annotations $B_{s}$. The feature alignment loss $\mathcal{L}_{adv}$ consists of four parts, which are MDQFA loss on the encoder, MTWFA loss on the encoder, MDQFA loss on the decoder and MTWFA on the decoder:
\begin{equation}
\fontsize{9}{10}\selectfont
\begin{aligned}
    \mathcal{L}_{adv} &= \sum_{l=1}^{L_{enc}}( \lambda_{\text{MDQFA}} \cdot \mathcal{L}^{\text{MDQFA}}_{enc_{l}} + \lambda_{\text{MTWFA}} \cdot \mathcal{L}^{\text{MTWFA}}_{enc_{l}}) +  \\
    &\sum_{l=1}^{L_{dec}}( \lambda_{\text{MDQFA}} \cdot \mathcal{L}^{\text{MDQFA}}_{dec_{l}} + \lambda_{\text{MTWFA}} \cdot \mathcal{L}^{\text{MTWFA}}_{dec_{l}})
\end{aligned}
\end{equation}
After obtaining a robust pretrained model in the pretraining stage, we then proceed to conduct self-training. The loss function of this stage $L_{st}$ can be formulated as:
\begin{equation}
\fontsize{9}{10}\selectfont
    \mathcal{L}_{st} = \mathcal{L}_{det}(I_{s},B_{s})+ \mathcal{L}_{det}(I_{t},\hat{B_{t}}) - \mathcal{L}_{adv}(I_{s}) - \mathcal{L}_{adv}(I_{t})
\end{equation}

Where $I_t$ represents target images and $\hat{B_{t}}$ refers to pseudo labels produced by the teacher model for $I_t$. To summarize, the overall training objective of MTM is defined as:
\begin{equation}
\fontsize{9}{10}\selectfont
    \underset{G}{\text{min}}~\underset{D}{\text{max}} \mathcal{L}_{det}(G) - \mathcal{L}_{adv}(G,D)
\end{equation}
where $G$ is the object detector and $D$ represents the domain discriminators.

\begin{table}[t]
\fontsize{9pt}{10pt}\selectfont
\centering
    \begin{tabular}{c|c}
        \hline
         Method & AP on car \\ 
         \hline
         Deformable DETR(source) & 47.4 \\
         SFA\cite{wang2021exploring} & 52.6 \\ 
         $\text{O}^{2}$Net\cite{gong2022improving} & 54.1 \\
         DA-DETR\cite{zhang2023detr} & 54.7 \\
         MTTrans\cite{yu2022cross} & 57.9 \\
         MTM(ours) & \textbf{58.1} \\
         \hline
    \end{tabular}
    \caption{Results(\%) in the synthetic to real adaptation scenario, \emph{i.e.} Sim10k $\rightarrow$ Cityscapes.}
    \label{tab:syn to real results}
\end{table}

\section{Experiment}
In this section, we conduct extensive experiments to 
testify our contributions: (1) pretraining with labeled target-like images avoids performance fluctuation (I1) and improves target performance; (2) OQKT helps the student model earn consistent performance gains in the self-training stage (I4); (3) our masked feature alignment methods including MDQFA and MTWFA prevent training stagnation (I2) and provide significant performance improvements in both stages.
\subsection{Datasets and Settings}
\subsubsection{Datasets}
We evaluate MTM on three challenging domain adaptation scenarios. \emph{i.e.} weather adaptation, scene adaptation, and synthetic to real adaptation. A detailed introduction of the three scenarios is as below:
\begin{itemize}
    \item{Weather Adaptation:}
    Cityscapes \cite{cordts2016cityscapes} is a landscape dataset containing 2975 training and 500 validation images. Foggy Cityscapes \cite{sakaridis2018semantic} is generated from Cityscapes by a fog synthesis algorithm. In this scenario, Cityscapes is the source dataset and Foggy Cityscapes is the target dataset.
    \item{Scene Adaptation:}
    In this scenario, Cityscapes also serves as the source dataset. BDD100K \cite{yu2020bdd100k} is an autonomous driving dataset consisting of 100k HD video clips. We utilize the \emph{daytime} subset of BDD100K which contains 36728 training images and 5258 validation images as the target dataset in this scenario.
    \item{Synthetic to Real Adaptation:}
    Sim10K \cite{johnsonroberson2017driving} is generated by the Grand Theft Auto game engine, containing 10,000 synthetic game images. In this scenario, Sim10K serves as the source dataset and Cityscapes serves as the target dataset.
\end{itemize}

\subsubsection{Comparative Benchmarks}
We compare MTM with five baselines built on Deformable DETR to validate the effectiveness of our proposed framework: Deformable DETR only trained on source data \cite{zhu2021deformable}, SFA \cite{wang2021exploring}, $\text{O}^2$Net \cite{gong2022improving}, MTTrans \cite{yu2022cross} and DA-DETR \cite{zhang2023detr}.

\subsubsection{Evaluation Metric}
We report the Average Precision on the car category with a threshold of 0.50 in the synthetic to real adaptation scenario. We adopt the mean Average Precision (mAP) of a threshold of 0.50 in the other two scenarios.

\subsubsection{Implementation Details}
We train CycleGAN \cite{zhu2017unpaired} for 100 epochs with a batch size of 8 and a learning rate of $2 \times 10^{-4}$ which linearly decreases to 0 after 50 epochs. We adopt an ImageNet \cite{deng2009imagenet}-pretrained ResNet50 network as the CNN backbone. In the pretraining stage of 80 epochs, we use an Adam optimizer with an initial learning rate $2 \times 10^{-4}$ decayed by 0.1 every 40 epochs. In the self-training stage of 20 epochs, the learning rate is $2 \times 10^{-6}$. The filtering threshold of pseudo labels is 0.50. Following \cite{wang2021exploring}, in the weather adaptation scenario, $\lambda_{\text{MTWFA}}$ is 1, and $\lambda_{\text{MDQFA}}$ is 0.1. In other scenarios, they are set to 0.01 and 0.001 respectively. $\theta_{Mask}$ and $\eta$ are set to 0.40 and 0.50. We use one 24GB GeForce RTX 3090 GPU in all experiments. 
Each batch includes 1 image from the source domain and 1 image from either the target-like domain or the target domain.

\subsection{Comparative Study}
The evaluations of our MTM are conducted on three domain adaptation scenarios. In the weather adaptation scenario (Table \ref{tab:weather results}), MTM outperforms the SOTA by 2.1 mAP. In the scene adaptation scenario (Table \ref{tab:scene results}), MTM shows a significant performance improvement of 4.7 mAP over the current SOTA method. In the synthetic to real adaptation scenario (Table \ref{tab:syn to real results}), MTM demonstrates a performance improvement of 0.2 mAP compared to MTTrans \cite{yu2022cross}, but still has a large improvement compared to the other baselines. 

\subsection{Ablation Studies}
Given that our framework is made up of two stages, the ablation studies are conducted in each stage in the weather adaptation scenario to prove the effectiveness of each component. Results of the ablation studies are presented in Table \ref{tab:pretrain ablation}-\ref{tab:sst ablation}.
\subsubsection{Pretraining Stage}
Several observations can be drawn from Table \ref{tab:pretrain ablation}: (1) pretraining with target-like images significantly enhances the model's target performance, resulting in a 13.8 mAP improvement; (2) MDQFA and MTWFA further improve performance compared to their non-masked counterparts. MDQFA and MTWFA further improve the model's target performance by 0.8 mAP and 0.9 mAP compared to their non-masked counterpart respectively. Combining both of them culminates in a 1.9 mAP improvement over their non-masked counterparts combination.

As depicted in Figure \ref{fig:performance curves}, we can observe that: (1) training with target-like images indeed avoids performance fluctuation (I1); (2) when conducting non-masked feature alignment on source images and target-like images, target performance plateaus after 40 epochs, indicating training stagnation (I2). Yet, masked feature alignment prevents this issue and maintains performance growth even after 40 epochs.

In Table \ref{tab:mask settings}, we present the target performance of pretrained models under different mask settings controlled by hyperparameters $\theta_{Mask}$ and $\eta$. The best pretraining performance is attained when $\theta_{Mask}=0.40$ and $\eta=0.50$. 
From Table \ref{tab:mask settings}, we find that setting the mask threshold $\theta_{Mask}$ of MDQFA between 0.30 and 0.50, and the mask threshold $\eta \cdot \theta_{Mask}$ of MTWFA between 0.10 and 0.30, further enhances the target performance of the pretrained model. The disparate optimal threshold ranges of the two methods are likely due to the fact that they handle features of different levels. MDQFA deals with global domain queries, so its domain discriminators are capable of distinguishing domains even when a significant portion of the information is masked. However, MTWFA handles local-level and instance-level sequence features, thus it will easily cause confusion to the domain discriminators of MTWFA if too much information is dropped from the mask.

\subsubsection{Self-training Stage}
Table \ref{tab:sst ablation} reveals notable observations as below: (1) MT helps increase target performance; (2) OQKT helps MT framework further improve target performance. With a pretrained model of 47.2 mAP, MT with OQKT outperforms standard MT by 0.2 mAP. MT achieves a further performance gain of 2.5 mAP if incorporated with OQKT; (3) MDQFA and MTWFA improve target performance by 0.1 mAP and 0.2 mAP respectively, and their combination brings a performance gain of 0.4 mAP. 

\begin{table}[t]
\centering 
\fontsize{9pt}{10pt}\selectfont
    \begin{tabular}{c c c c c}
        \hline
        {Method} & {Target-like} & {MTWFA}& {MDQFA} & {mAP} \\  
        \hline
        \multirow{2}{*}{D-DETR} &  &  &   & 28.5 \\
        \multirow{2}{*}{} & \ding{51} &  &  & 42.3 \\
        \hline
        \multirow{6}{*}{Proposed} & \ding{51} & $\bigcirc$ &  & 42.7 \\
        \multirow{6}{*}{} & \ding{51} & \ding{51} &  & 43.6 \\
        \multirow{6}{*}{} & \ding{51} &  & $\bigcirc$ & 44.9 \\
        \multirow{6}{*}{} & \ding{51} &  & \ding{51} & 45.7 \\
        \multirow{6}{*}{} & \ding{51} & $\bigcirc$ & $\bigcirc$ & 45.3 \\
        \multirow{6}{*}{} & \ding{51} & \ding{51} &  \ding{51} & \textbf{47.2} \\
        \hline
    \end{tabular}
    \caption{Ablation study of the pretraining stage in the weather adaptation scenario. D-DETR stands for Deformable-DETR trained on source data. $\bigcirc$ refers to adopting the non-masked feature alignment method. \\}
    \label{tab:pretrain ablation}
    \begin{tabular}{|c | c | c | c | c | c | c | c | c |}
        \hline
        \multicolumn{2}{|c|}{$\theta_{Mask}$} & 0.20 & 0.30 & 0.40 & 0.50 & 0.60 & 0.70 & 0.80 \\ 
        \hline
        \multirow{3}{*}{$\eta$} & 1/3 & 45.3 & 45.5 & 46.1 & \textbf{46.8} & 45.2 & 45.0 & 44.5  \\
        \multirow{3}{*}{} & 1/2 & 45.6 & 46.2 & \textbf{47.2} & 45.8 & 44.9 & 44.3 & 44.2  \\
        \multirow{3}{*}{} & 1 & 45.6 & 45.8 & \textbf{46.6} & 45.5 & 44.9 & 44.6 & 43.4 \\
        \hline
    \end{tabular}
    \caption{Performance of pretrained models in the weather adaptation scenario under different mask settings. \\}
    \label{tab:mask settings}

    \begin{tabular}{c c c c c c}
        \hline
        Method & MT & OQKT & MTWFA & MDQFA & mAP \\
        \hline
        \multirow{3}{*}{D-DETR} &  &  &  &  & 28.5 \\
        \multirow{3}{*}{} & \ding{51} &  &  &  & 35.8 \\
        \multirow{3}{*}{} & \ding{51} & \ding{51} &  & & 38.3
        \\
        \hline
        \multirow{6}{*}{Proposed} & & & & & 47.2 \\
        \multirow{6}{*}{} & \ding{51} & & & & 47.7 \\
        \multirow{6}{*}{} & \ding{51} & \ding{51} & & & 47.9 \\
        \multirow{6}{*}{} & \ding{51} & \ding{51} & & \ding{51} & 48.0 \\
        \multirow{6}{*}{} & \ding{51} & \ding{51} & \ding{51} & & 48.1 \\
        \multirow{6}{*}{} & \ding{51} & & \ding{51} & \ding{51} & 48.4 \\
        \multirow{6}{*}{} & \ding{51} & \ding{51} & \ding{51} & \ding{51} & \textbf{48.9} \\    
        \hline
    \end{tabular}
    \caption{Ablation study of the self-training stage in the weather adaptation scenario. MT refers to utilizing the mean teacher framework for self-training.}
    \label{tab:sst ablation}
\end{table}

\begin{table}[h]
\centering
\fontsize{9pt}{10pt}\selectfont
\end{table}

\section{Theoretical Analysis}
This section analyzes our framework from a theoretical aspect based on \cite{blitzer2007learning,ben2010theory}.

Let $\mathcal{H}$ be a hypothesis space of VC dimension $d$. For each $i \in \{1,...,N\}$, let $S_i$ be a labeled sample of size $\beta_j m$ generated by drawing $\beta_j m$ points from domain $\mathcal{D}_i$ and labeling them according to $f_i$, and $\alpha_i$ represents domain weight of $\mathcal{D}_i$. $N$ is set to 2 in our work because we only use two types of labeled images including source images and target-like images. If $\hat{h} \in \mathcal{H}$ is an empirical hypothesis of the empirical $\alpha$-weighted error $\hat{\epsilon}_{\alpha}(h)$ on these multi-source samples, we have the following theorem to bound the target error of the empirical hypothesis $\hat{h}$:

\newtheorem{theorem}{Theorem}
\begin{theorem}
For any $\delta \in (0,1)$, with probability $1-\delta$,
\fontsize{9}{10}\selectfont
\begin{equation}
\begin{aligned}
    \epsilon_{T}(\hat{h}) &\leq \hat{\epsilon}_{\alpha}(\hat{h}) + \frac{1}{2}d_{\mathcal{H}\Delta \mathcal{H}}(D_{\alpha},D_T) + \gamma_\alpha \\
    &+ \sqrt{(\sum^{2}_{j=1}\frac{\alpha_j^2}{\beta_j})(\frac{dln(2m)-ln(\delta)}{2m})} \\
\end{aligned}    
\end{equation}
\label{theorem: error inequality}
\end{theorem}
where $\gamma_\alpha=min_h\{\epsilon_T(h)+\epsilon_\alpha(h)\}$ represents the error of the joint ideal hypothesis that is correlated with the ability of the detector and the discriminability of features. $d_{\mathcal{H}\Delta \mathcal{H}}(D_{\alpha},D_T)$ is the domain divergence that is associated with the domain-invariance of features. 
% Please refer to the Appendix for detailed proof of Theorem \ref{theorem: error inequality}. 

In Theorem \ref{theorem: error inequality}, the target error $\epsilon_{T}(\hat{h})$ can be bounded by four factors: (1) the empirical $\alpha$-weighted error on the multi-source samples $\hat{\epsilon}_{\alpha}(\hat{h})$, which can be minimized by applying supervised loss on the multi-source samples; (2) the domain divergence between the multi-source and the target domain $d_{\mathcal{H}\Delta \mathcal{H}}(D_{\alpha},D_T)$. Our masked feature alignment methods alleviate domain shift to minimize the domain divergence; (3) the error of the joint ideal hypothesis $\gamma_{\alpha}$. Both our pretraining stage and self-training stage help enhance the model's ability to reduce this value; (4) a complexity term whose minimum value is obtained when $\forall j \in \{1,2\}, \alpha_j = \beta_j$. This can be achieved by assigning equal weight to each sample point in the multi-source data. Based on the analysis above, we can minimize the target error $\epsilon_{T}(\hat{h})$ by MTM.

\section{Conclusion}
In this work, we deeply investigate UDAOD methods based on DETR and uncover several issues. Previous one-stage feature alignment methods overlook their inherent issues: performance fluctuation and training stagnation, while the two-stage feature alignment method based on mean teacher introduces new challenges like unreliable pretrained model and unstable performance gains. Besides, how to utilize the third related domain such as the target-like domain to assist domain adaptation remains unexplored in existing methods.
To address the issues and build a robust domain adaptive Detection Transformer framework, we propose a two-stage framework named MTM, \emph{i.e.} Mean Teacher DETR with Masked Feature Alignment.

In the pretraining stage, CycleGAN is used to generate target-like images. We incorporate generated target-like images in pretraining to avoid performance fluctuation. In the self-training stage, we leverage unlabeled target images by pseudo labels based on mean teacher. We propose Object Queries Knowledge Transfer (OQKT) to achieve consistent performance gains. Above all, we propose masked feature alignment including Masked Domain Query-based Feature Alignment (MDQFA) and Masked Token-Wise Feature Alignment (MTWFA) to alleviate domain shift in a more robust way. Our masked feature alignment methods not only prevent training stagnation and lead to a robust pretrained model in the pretraining stage but also enhance the model's final target performance in the self-training stage.

Experimental results and a theoretical analysis have proven the effectiveness of MTM. We expect that our research will inspire future work in this area.

\section{Acknowledgements}
This work was supported by the National Key R\&D Program of China (2022YFB4701400/4701402), SSTIC Grant(KJZD20230923115106012), Shenzhen Key Laboratory (ZDSYS20210623092001004), and Beijing Key Lab of Networked Multimedia.

\nocite{*}

\bibliography{aaai24}

\end{document}